\documentclass{article}
\usepackage{spconf,amsmath,graphicx,booktabs,tabularx,url}
\usepackage{multirow}
\usepackage[ruled,vlined]{algorithm2e}
\usepackage{placeins}
\usepackage{CJKutf8}

\newcommand{\method}{\textsc{Smart}}

\title{SMART: SEMANTIC SEGMENT MAPPING AND ASSESSMENT OF RECOGNITION TRANSCRIPTS WITH CONTEXTUAL RUBRICS}

\name{Xiaoyu Gu$^{1}$\sthanks{denotes equal contribution.}, Chencheng Wang$^{1,*}$, Fuyang Li$^{2}$, Yu Xi$^{1}$, Tao Liu$^{3}$, Jiaoqi Guo$^{3}$, Kai Yu$^{1}$\sthanks{denotes the corresponding author.}}
\address{$^{1}$MoE Key Lab of Artificial Intelligence, AI Institute, X-LANCE Lab, Shanghai Jiao Tong University\\
$^{2}$School of Intelligence Science and Technology, Nanjing University, Suzhou, China\\
$^{3}$AISpeech Ltd, Suzhou, China}

\begin{document}
\ninept
\maketitle

\begin{abstract}
Word error rate (WER) assigns equal cost to lexical edits despite
their different semantic consequences, while overall semantic
scores cannot identify which individual errors alter the intended
meaning. We formulate ASR semantic evaluation as context-aware,
segment-level consequence assessment and propose Semantic Segment
Mapping and Assessment of Recognition Transcripts with Contextual
Rubrics (\method{}), a framework for traceable semantic evaluation
and targeted ASR correction. \method{} maps reference--hypothesis
discrepancies into coherent segments and applies a structured
contextual rubric to assign explicit consequence labels. These labels
support Semantic Impact Error Rate (SIER), which quantifies
consequential errors at utterance, chunk, and lexical levels.
Against majority-vote human labels, \method{} achieves 0.843 F1
versus 0.745 for direct LLM judgment. Evaluation across multiple
LLMs further demonstrates that \method{} is relatively insensitive
to the choice of evaluation model, while MER and SIER reveal
different aspects of ASR quality. A separate target-reference-free
rubric guides correction using the target hypothesis and
surrounding ASR hypotheses, outperforming a matched
\emph{w/o Rubric} baseline on both datasets, with further
absolute reductions of 0.85--0.98 percentage points in MER
and 1.18--1.65 percentage points in
$\mathrm{SIER}_{\mathrm{lex}}$.
\end{abstract}

\vspace{-5pt}

\section{Introduction}
\label{sec:intro}
\vspace{-5pt}

Automatic speech recognition (ASR) errors can have markedly different
semantic consequences. Omitting a negation may reverse the intended
meaning, whereas inserting a filler may leave it unchanged. Crucially,
errors with different consequences can coexist within the same
utterance. Word error rate (WER) assigns equal cost to lexical
substitutions, deletions, and insertions, regardless of their impact
on understanding
\cite{wang_wer_slu,he_wer_translation,favre_human_utility}.
Semantic ASR evaluation must therefore go beyond overall meaning
preservation to identify which individual recognition errors actually
alter the intended meaning.

Existing approaches address the limitations of WER from different
perspectives. Downstream task-based evaluation measures the impact of
ASR errors through tasks such as intent recognition, slot filling,
and semantic parsing
\cite{wang_wer_slu,he_wer_translation,pulikodan_aer}.
However, its scope is determined by task-specific information
requirements: semantic distortions that do not affect the task
outcome may go undetected. Consequently, such evaluation cannot
comprehensively assess the preservation of fine-grained semantic
information or systematically identify the consequences of
individual recognition errors.

Sentence-level semantic evaluation instead measures the overall
agreement between reference and hypothesis using contextualized
representations and semantic similarity
\cite{zhang_bertscore,zhao_moverscore,kim2021semdist,
kim2022perception,rugayan2022asd},
or LLM-based judgments
\cite{wang_interactive_asr,ellis_wer_unaware}.
By reducing multiple discrepancies to an overall assessment, these
approaches do not explicitly distinguish errors that alter the
intended meaning from those that leave it intact, limiting error
localization and fine-grained analysis.

Hybrid and severity-aware metrics further distinguish recognition
errors through lexical--semantic weighting or error severity
\cite{sasindran_heval,whetten_severity}.
Localized approaches introduce segment-wise alignment and semantic
scoring \cite{sasindran2024semascore}, while LLM-based methods
assess recognition errors through structured penalties
\cite{parulekar_laser}.
Nevertheless, assigning scores or penalties to local discrepancies
does not directly establish whether each segment causes a semantic
consequence in context.

Discourse context is essential to segment-level consequence
assessment. Isolated reference--hypothesis spans may be insufficient
to resolve anaphora, ellipsis, or dependencies across utterances.
Recent studies have also examined how ASR errors affect understanding
in task-specific or dialogue contexts
\cite{wang_interactive_asr,ellis_wer_unaware}.
Yet existing localized scoring approaches primarily focus on the
target utterance or its constituent spans, without explicitly
assessing the semantic consequence of each segment in its
surrounding discourse context.

We therefore formulate semantic ASR evaluation as a
\emph{context-aware, segment-level, and traceable semantic
consequence assessment}: reference--hypothesis discrepancies are
organized into semantically coherent segments, each of which is
evaluated in discourse context and linked to its corresponding
recognition errors.

To address this task, we propose
\textbf{S}emantic Segment \textbf{M}apping and
\textbf{A}ssessment of \textbf{R}ecognition
\textbf{T}ranscripts with Contextual Rubrics
(\method{}), a framework for semantic consequence evaluation
and ASR correction. Its evaluation branch constructs coherent
reference--hypothesis segments through character-level alignment
and language-aware boundaries. A structured rubric then assesses
each discrepant segment using surrounding reference context,
with deterministic aggregation producing explicit consequence
labels. These labels support Semantic Impact Error Rate (SIER),
which quantifies consequential errors at utterance, chunk, and
lexical levels. A separate target-reference-free correction branch
uses the target hypothesis and surrounding ASR hypotheses to
generate structured repair guidance.

Our main contributions are:
\vspace{-5pt}

\begin{enumerate}
\item \textbf{Semantically coherent segment mapping.}
We combine character-level alignment with language-aware boundaries
to construct coherent reference--hypothesis segments, enabling
semantic consequences to be localized and traced to specific
recognition errors.

\begin{figure*}[t]
    \centering
    \includegraphics[width=0.96\textwidth]{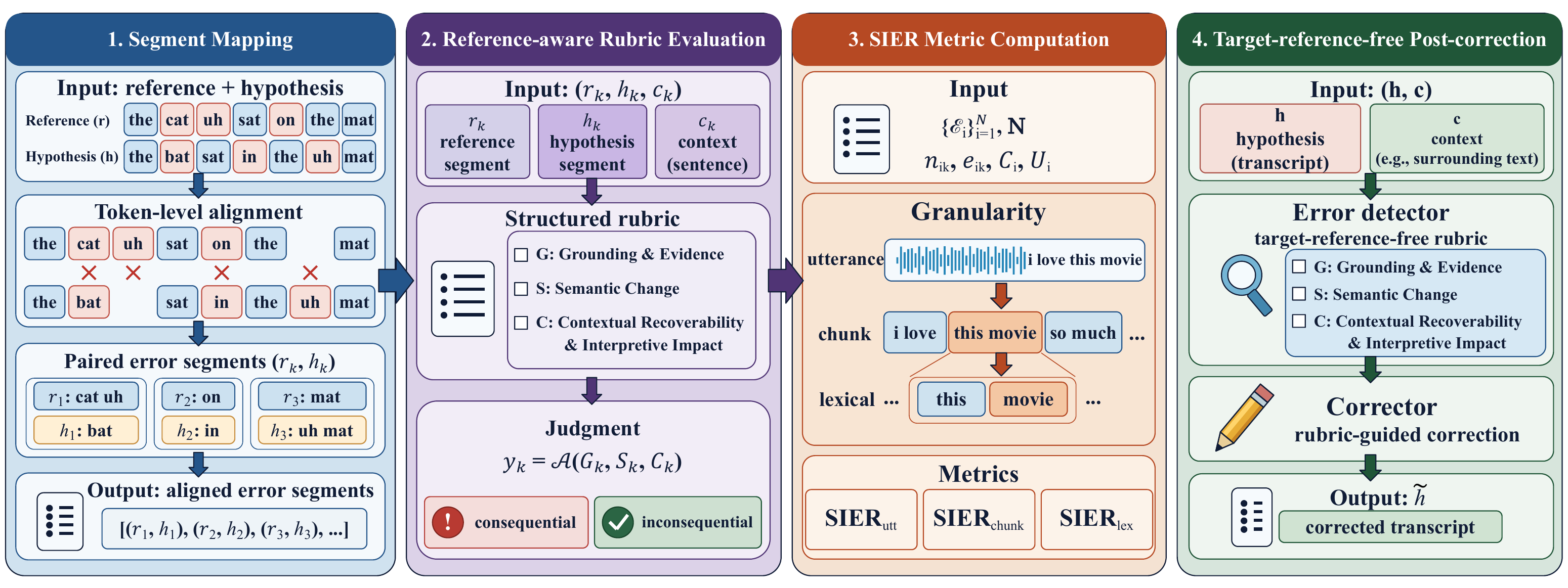}

    \caption{\textbf{SMART: Context-aware segment-level semantic
    consequence assessment and correction.}
    \textbf{(1) Segment Mapping:} Organizes aligned lexical
    discrepancies into semantically coherent segments.
    \textbf{(2) Reference-aware Rubric Evaluation:} Judges the
    semantic consequence of each segment using surrounding
    reference context and structured criteria.
    \textbf{(3) SIER Metric Computation:} Quantifies consequential
    errors at utterance, chunk, and lexical levels.
    \textbf{(4) Target-reference-free Post-correction:} Uses the
    target hypothesis and surrounding ASR hypotheses to identify
    potential semantic errors and guide LLM repair.}
    \label{fig:framework}
\end{figure*}

\item \textbf{Context-aware consequence assessment and SIER.}
We introduce a structured contextual rubric with deterministic
aggregation to assign explicit consequence labels to individual
segments. The resulting SIER quantifies semantic-impact errors
at utterance, chunk, and lexical levels, enabling fine-grained
error analysis and direct comparison with MER at the lexical level.

\item \textbf{Target-reference-free rubric-guided correction.}
We develop a separate correction rubric that guides targeted
LLM repair using only the target hypothesis and surrounding ASR
hypotheses, extending structured semantic analysis to correction
without access to reference transcripts.
\vspace{-10pt}
\end{enumerate}

\section{SMART Framework}
\label{sec:methodology}
\vspace{-5pt}

\subsection{Framework Overview}
\label{sec:overview}

As shown in Fig.~\ref{fig:framework}, Semantic Segment Mapping and Assessment of Recognition
Transcripts with Contextual Rubrics (\method{}) contains a reference-aware
evaluation branch and a target-reference-free correction branch. For
evaluation, the reference $r$ and hypothesis $h$ are mapped into aligned units
$\mathcal{M}=\{(r_k,h_k)\}_{k=1}^{K}$ with discrepancy indices
$\mathcal{I}$. Each discrepant unit is assessed with surrounding reference
context $c^{\mathrm{ref}}$ by a structured rubric and deterministic aggregation
for SIER. For correction, a separate rubric uses only the target hypothesis $h$
and surrounding ASR hypotheses $c^{\mathrm{hyp}}$ to produce repair guidance
$\mathcal{D}^{\mathrm{rep}}$, from which an LLM generates $\tilde{h}$.
Reference-aware evaluation diagnoses are not passed to correction; the revised
hypothesis is evaluated afterward using MER and SIER.
\vspace{-5pt}

\begin{algorithm}[h]
\caption{Semantic Segment Mapping}
\label{alg:segment-mapping}
\DontPrintSemicolon
\KwIn{Reference $r$; hypothesis $h$.}
\KwOut{Segments $\mathcal{M}$; discrepancy indices $\mathcal{I}$.}

Normalize $r,h$ and compute character-level Levenshtein alignment.\;
Extract chunk-start boundaries for English whitespace tokens or Chinese jieba
tokens.\;
Initialize segment list $U$ and the current aligned pair.\;

\ForEach{alignment column}{
    \If{aligned spaces or both positions reach chunk starts}{
        Append the current pair to $U$ and reset it.\;
    }
    Append non-gap characters to the current pair and advance positions.\;
}
Append the remaining pair and remove empty pairs.\;
Merge one-sided empty pairs with the next segment, or the previous one at the
end.\;
$\mathcal{M}\leftarrow[(r_k,h_k)]_{k=1}^{K}$ from $U$;\quad
$\mathcal{I}\leftarrow\{k\mid r_k\neq h_k\}$.\;
\Return $(\mathcal{M},\mathcal{I})$.\;
\end{algorithm}
\vspace{-5pt}

\subsection{Semantically Coherent Segment Mapping}
\label{sec:segment_mapping}
\vspace{-5pt}
Lexical edits may fragment one semantic discrepancy, especially under
insertions, deletions, and word splits or merges. We therefore construct local
evaluation segments from character-level Levenshtein alignment\cite{levenshtein1966,wagner1974} and
language-aware chunk boundaries. Chunks are whitespace-delimited English
tokens or jieba-tokenized Chinese units; segments are aligned
reference--hypothesis pairs used for semantic judgment. A segment is split
only at shared chunk boundaries, which keeps unequal split/merge errors
together. One-sided empty segments are attached to a near segment so
that isolated insertions and deletions retain local context.
Algorithm~\ref{alg:segment-mapping} returns
\begin{equation}
\mathcal{M}=[(r_k,h_k)]_{k=1}^{K},\qquad
\mathcal{I}=\{k\mid r_k\neq h_k\},
\label{eq:segment_mapping}
\end{equation}
$\mathcal{M}$ contains aligned evaluation segments and $\mathcal{I}$
indexes discrepancies passed to semantic consequence evaluation.
\vspace{-5pt}

\subsection{Reference-Aware Contextual Semantic Consequence Judgment}
\label{sec:rubric}
\vspace{-5pt}
For each discrepant segment $k\in\mathcal{I}$, the evaluation rubric receives
the aligned pair $(r_k,h_k)$ with surrounding reference context $c_k^{\mathrm{ref}}$.
Inspired by structured LLM-based evaluation
\cite{parulekar_laser,liu_geval,hashemi_llm_rubric},
we elicit a rubric trace rather than directly predicting
a consequence label:
\begin{equation}
z_k=(G_k,S_k,C_k),
\label{eq:rubric-trace}
\end{equation}
where $G_k$ evaluates alignment validity and evidence sufficiency, $S_k$ characterizes the semantic change between reference and hypothesis, and $C_k$ assesses whether the reference-intended meaning remains recoverable under the surrounding reference context and how the discrepancy affects interpretation. The structured trace is converted into a segment-level label by fixed
deterministic rules,
\vspace{-5pt}

\begin{equation}
\begin{aligned}
y_k&=\mathcal{A}(G_k,S_k,C_k),\\
y_k&\in\{\textsc{Consequential},\textsc{Inconsequential}\}.
\end{aligned}
\label{eq:rubric-aggregation}
\end{equation}
Conceptually, a discrepancy is consequential when its semantic change remains meaningfully disruptive after considering contextual recoverability and interpretive impact. The resulting labels define the consequential segments
used to compute SIER.
\vspace{-5pt}
\subsection{Target-Reference-Free Rubric-Guided Correction}
\label{sec:correction}
\vspace{-5pt}

We extend the same structured reasoning principles to correction through a
separate rubric that operates only on the target hypothesis $h$ and surrounding
ASR hypotheses $c^{\mathrm{hyp}}$, without access to reference transcripts. It identifies potentially
problematic regions and produces repair-oriented guidance
$\mathcal{D}^{\mathrm{rep}}$, which is then used to revise the hypothesis:
\begin{equation}
\begin{aligned}
\mathcal{D}^{\mathrm{rep}}&=\mathcal{R}_{\mathrm{rep}}(h,c^{\mathrm{hyp}}),\\
\tilde{h}&=\mathcal{F}
\bigl(h,c^{\mathrm{hyp}},\mathcal{D}^{\mathrm{rep}};\pi_{\mathrm{repair}}\bigr).
\end{aligned}
\label{eq:diagnosis-guided-correction}
\end{equation}
Here, $\mathcal{F}$ denotes the prompted correction LLM \cite{ma_asr_correction,li_multilingual_correction,madaan_self_refine}. The correction branch
receives neither reference transcripts nor reference-aware evaluation
diagnoses. The revised hypothesis $\tilde{h}$ is evaluated afterward using
MER and SIER.
\vspace{-10pt}

\section{Experimental Setup}
\label{sec:experimental-setup}
\vspace{-5pt}

\subsection{Datasets}
\label{sec:datasets}
\vspace{-5pt}

We evaluate \method{} on two continuous-speech datasets covering
different languages and speaking domains. Both predominantly feature
a single speaker, allowing discourse context to be examined without
multi-speaker interaction effects.
\begin{itemize}
\item \textbf{Canvas Speech}: An internal Mandarin classroom dataset
comprising 3.4 hours of continuous instructional recordings with
reference transcripts. The recordings contain Mandarin--English
code-switching and domain-specific terminology, providing a realistic
educational setting for multilingual ASR evaluation.

\item \textbf{TED-LIUM release 3}~\cite{hernandez2018tedlium3}:
A publicly available English speech corpus containing TED-talk recordings
and reference transcripts. We select a 7.2-hour subset of continuous
presentations to evaluate semantic consequences in long-form English speech.
\end{itemize}

For both datasets, evaluation context $c^{\mathrm{ref}}$ comprises the
ten consecutive ground-truth utterances immediately before and after
each target utterance. Correction uses the same temporal window of
corresponding ASR hypotheses $c^{\mathrm{hyp}}$, without access to
reference transcripts.
\vspace{-5pt}
\subsection{Models and Evaluation Metrics}
\label{sec:models-metrics}
\vspace{-5pt}
DeepSeek-V4-Flash is used for the main evaluation and correction
experiments. Additional evaluation LLMs are examined in the
sensitivity analysis.
For human evaluation, we sample 300 discrepant segments per dataset, with
approximately 30\% consequential cases. Five in-domain annotators label each
segment independently, and majority vote defines the ground truth. Rubrics and
aggregation rules are frozen before evaluation. We report accuracy, precision,
recall, and F1; BERTScore and SeMaScore use oracle evaluation-set thresholds
that maximize F1. We use \emph{Mixed Error Rate} (MER) as the lexical-error baseline,
treating Chinese characters and English words as lexical units:
\begin{equation}
\mathrm{MER}=\frac{\sum_i(S_i+D_i+I_i)}{\sum_i L_i},
\label{eq:mer}
\end{equation}
where $S_i,D_i,I_i$ are substitution, deletion, and insertion counts
and $L_i$ is the reference mixed-unit count. MER reduces to CER or
WER for Chinese-only or English-only speech, respectively.

Our main semantic metric, \emph{Semantic Impact Error Rate} (SIER),
measures error mass within consequential segments. Let
$\mathcal{E}_i$ denote the consequential segments in utterance $i$:
\begin{equation}
\begin{aligned}
\mathrm{SIER}_{\mathrm{utt}}
&= \frac{\sum_i\mathbf{1}[|\mathcal{E}_i|>0]}{N}, \\
\mathrm{SIER}_{\mathrm{chunk}}
&= \frac{\sum_i\sum_{k\in\mathcal{E}_i}n_{ik}}{\sum_i C_i}, \\
\mathrm{SIER}_{\mathrm{lex}}
&= \frac{\sum_i\sum_{k\in\mathcal{E}_i}e_{ik}}{\sum_i L_i}.
\end{aligned}
\label{eq:sier}
\end{equation}
Here, $n_{ik}$ and $e_{ik}$ are the hypothesis-side chunk count
and MER-style lexical edits in consequential segment $k$,
respectively; $C_i$ and $L_i$ denote the reference chunk and
mixed-unit counts. English whitespace and Chinese jieba tokens
define chunk boundaries.

Crucially, $\mathrm{SIER}_{\mathrm{lex}}$ uses the same lexical
edit counts and reference denominator as MER, but retains only
edits within consequential segments. The two metrics are thus
directly comparable: their difference measures the lexical error
mass excluded from SIER because it falls within segments judged
semantically inconsequential.

\textbf{Complete rubrics, prompts, aggregation rules, and code are
publicly available.}
\footnote{\url{https://github.com/NytePlus/SMART}}
\vspace{-5pt}
\section{Results and Analysis}
\label{sec:results}
\vspace{-5pt}
\subsection{Lexical and Semantic ASR Evaluation}
\label{sec:metric-comparison}
\vspace{-15pt}

\begin{table}[h]
\caption{\textbf{ASR evaluation on lexical and semantic.}
Results on Canvas Speech and TED-LIUM 3 datasets. MER uses Chinese characters
and English words as lexical units. All values are percentages;
bold indicates the best result within each model group.}
\label{tab:large-scale-evaluation}
\centering
\footnotesize
\setlength{\arrayrulewidth}{0.3pt}
\small
\setlength{\tabcolsep}{2.5pt}
\resizebox{\columnwidth}{!}{%
\begin{tabular}{lcc|cc|c}
\hline
\multirow{2}{*}{ASR System}
& \multicolumn{2}{c|}{Utterance-level}
& \multicolumn{3}{c}{Lexical / chunk-level} \\
\cline{2-6}
& SER $\downarrow$
& $\mathrm{SIER}_{\mathrm{utt}}$ $\downarrow$
& MER $\downarrow$
& $\mathrm{SIER}_{\mathrm{lex}}$ $\downarrow$
& $\mathrm{SIER}_{\mathrm{chunk}}$ $\downarrow$ \\
\hline
\multicolumn{6}{l}{\small\textbf{Open-Source Models}} \\
Qwen3-ASR\cite{shi2026qwen3asr}
& 98.07 & 80.94 & 12.97 & 6.08 & 6.40 \\
whisper-v3\cite{radford_whisper}
& \textbf{94.22} & \textbf{75.59} & 24.41 & 16.05 & 15.16 \\
whisper-v3-turbo\cite{radford_whisper}
& 95.29 & 76.66 & 23.19 & 15.16 & 14.34 \\
sensevoice\cite{an2024funaudiollm}
& 98.29 & 82.66 & \textbf{11.87} & \textbf{5.43} & \textbf{6.23} \\
paraformer\cite{gao2022paraformer}
& 98.07 & 84.15 & 15.04 & 6.09 & 7.22 \\
\hline
\multicolumn{6}{l}{\small\textbf{Closed-Source Models}} \\
Qwen3-ASR-Filetrans
& \textbf{84.15} & \textbf{56.75}
& \textbf{9.64} & 3.60 & \textbf{3.82} \\
Doubao Filetrans
& 87.79 & 62.10 & 11.13 & \textbf{3.42} & 4.03 \\
\hline
\end{tabular}
}
\end{table}
\vspace{-10pt}

As shown in Table~\ref{tab:large-scale-evaluation},
$\mathrm{SIER}_{\mathrm{utt}}$ is consistently lower than SER
across all ASR systems, indicating that some utterances contain
lexical errors without consequential semantic changes.
Moreover, Qwen3-ASR-Filetrans achieves lower MER than Doubao
Filetrans (9.64\% vs.\ 11.13\%) but higher
$\mathrm{SIER}_{\mathrm{lex}}$ (3.60\% vs.\ 3.42\%),
demonstrating that lexical accuracy and semantic impact can
yield different system rankings. Table~\ref{tab:metric-reversal-example} further illustrates this
discrepancy: WER favors hypothesis A despite its consequential
negation error, whereas $\mathrm{SIER}_{\mathrm{lex}}$ favors B
despite its additional lexical edits. Together, these results
show that SIER complements conventional error rates by
distinguishing semantic impact from lexical error frequency.
\vspace{-15pt}

\begin{table}[h]
\caption{\textbf{An example of differing lexical and semantic errors.}
Hypotheses A and B are produced by Qwen3-ASR-Filetrans
and Doubao Filetrans, respectively, for the same utterance
from TED-LIUM release 3. A omits the consequential negation
``not,'' whereas B preserves it but introduces inconsequential
fillers. WER favors A, while
$\mathrm{SIER}_{\mathrm{lex}}$ favors B.
All values are percentages; lower is better.}
\label{tab:metric-reversal-example}
\centering
\small
\setlength{\tabcolsep}{5pt}
\renewcommand{\arraystretch}{1.15}

\begin{tabularx}{\columnwidth}{
@{}>{\raggedright\arraybackslash}Xcc@{}
}
\toprule
\multicolumn{3}{@{}p{\columnwidth}@{}}{
\textbf{Context}\quad
Why can a longer vacation feel only slightly better?
}\\

\multicolumn{3}{@{}p{\columnwidth}@{}}{
\resizebox{\columnwidth}{!}{%
\textbf{Reference}\quad
There are no new memories added; you have \textbf{not}
changed the story.}
}\\

\hline
Hypothesis
& WER $\downarrow$
& $\mathrm{SIER}_{\mathrm{lex}}$ $\downarrow$ \\
\hline

\textbf{A:} There are no new memories added;
& \multirow[c]{2}{*}{\textbf{8.33}}
& \multirow[c]{2}{*}{8.33} \\

you have changed the story.
& & \\

\textbf{B:} Uh, well, there are no new memories
& \multirow[c]{2}{*}{16.67}
& \multirow[c]{2}{*}{\textbf{0.00}} \\

added; you have not changed the story.
& & \\

\hline
\end{tabularx}
\end{table}
\vspace{-15pt}
\subsection{Agreement with Human Judgments}
\label{sec:human-agreement}

As shown in Table~\ref{tab:semantic-detection}, BERTScore
and SeMaScore achieve high recall but low precision and accuracy.
Their similarity-based scoring may penalize lexical discrepancies
that do not alter the intended meaning, leading to excessive
consequential-error predictions. Interactive ASR exhibits a
similar pattern, with the highest recall (0.955) but low
precision (0.297), suggesting a conservative tendency in
judging meaning preservation.

Compared with Direct LLM, SMART w/o context achieves higher
accuracy (0.920 vs.\ 0.900) and F1 (0.822 vs.\ 0.745),
demonstrating the benefit of structured rubric assessment
even without discourse context. Incorporating reference
context further improves precision from 0.822 to 0.857
and recall from 0.822 to 0.830, yielding an F1 of 0.843.
These results highlight the contributions of both structured
segment-level judgment and discourse context.
\vspace{-10pt}
\begin{table}[h]
\caption{\textbf{Agreement with human semantic consequence judgments.}
Classification results on discrepant segments from Canvas
Speech and TED-LIUM 3, using majority-vote human annotations
as ground truth. \textsc{Consequential} is the positive class.
Acc., Prec., Rec., and F1 denote accuracy, precision, recall,
and F1-score, respectively. All LLM-based methods use
DeepSeek-V4-Flash. Bold indicates the highest value in
each column.}
\label{tab:semantic-detection}
\centering
\footnotesize
\small
\setlength{\tabcolsep}{2pt}
\begin{tabular*}{\columnwidth}{@{\extracolsep{\fill}}lrrrr@{}}
\hline
\textbf{Classifier}
& \textbf{Acc. $\uparrow$}
& \textbf{Prec. $\uparrow$}
& \textbf{Rec.} $\uparrow$
& \textbf{F1} $\uparrow$\\
\hline
\multicolumn{5}{l}{\textbf{Similarity-based metrics}} \\
BERTScore~\cite{zhang_bertscore}
& 0.215 & 0.214 & 0.910 & 0.331\\
SeMaScore~\cite{sasindran2024semascore}
& 0.370 & 0.261 & 0.910 & 0.384\\
\hline
\multicolumn{5}{l}{\textbf{LLM-based evaluation}} \\
Interactive ASR~\cite{wang_interactive_asr}
& 0.515 & 0.297 & \textbf{0.955} & 0.440\\
Direct LLM
& 0.900 & 0.764 & 0.731 & 0.745\\
SMART w/o context
& 0.920 & 0.822 & 0.822 & 0.822\\
SMART
& \textbf{0.925} & \textbf{0.857} & 0.830 & \textbf{0.843}\\
\hline
\end{tabular*}
\end{table}
\vspace{-15pt}

\subsection{Effectiveness of Target-Reference-Free Rubric-Guided Correction}
\label{sec:correction-effectiveness}

\begin{figure}[h]
    \centering
    \includegraphics[width=0.5\textwidth]{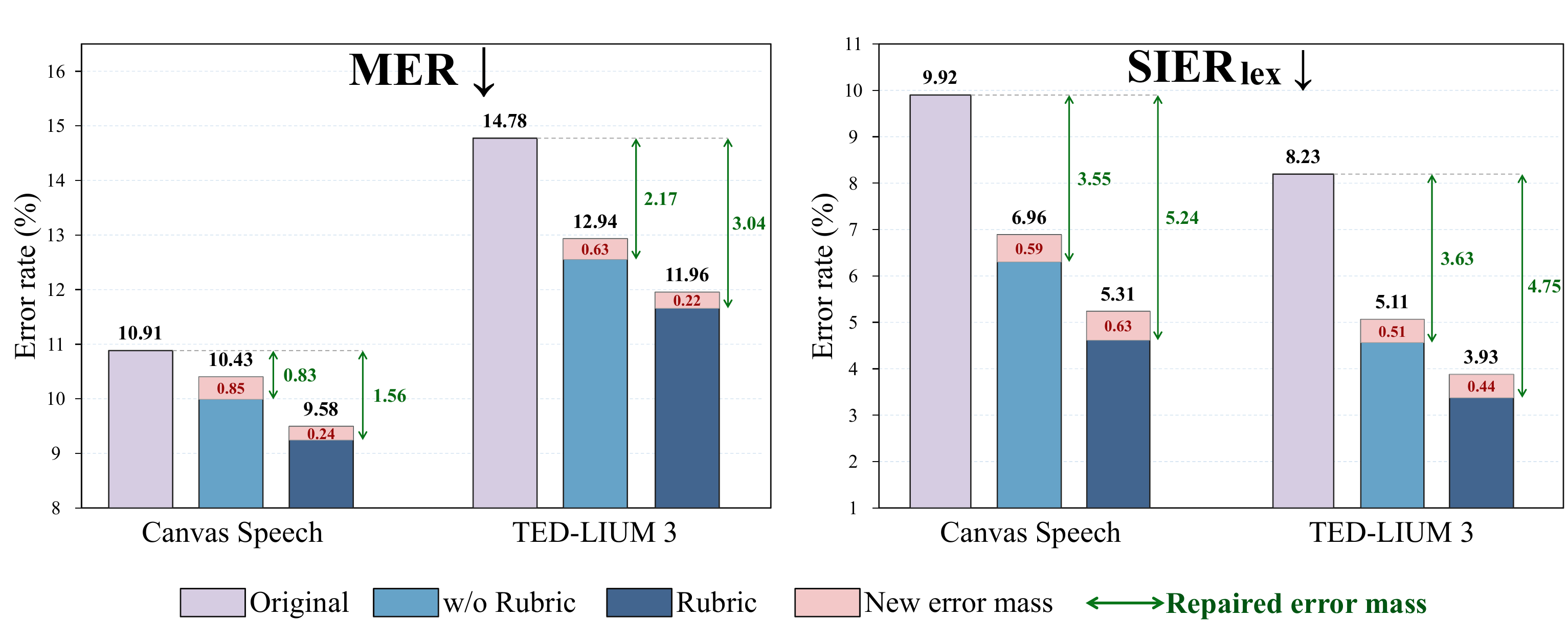}
    \caption{Target-reference-free correction of SenseVoice on Canvas Speech and
TED-LIUM 3. Bars report MER and $\mathrm{SIER}_{\mathrm{lex}}$ for the
original hypothesis, the matched \emph{w/o Rubric} baseline, and
rubric-guided correction. Pink caps indicate newly introduced error mass,
while double-headed arrows indicate repaired error mass relative to the
original hypothesis.}
    \label{fig:correction}
\end{figure}

We compare rubric-guided correction with a matched \emph{w/o Rubric}
baseline using the same target hypothesis, surrounding ASR hypotheses, model,
and decoding. As shown in
Fig.~\ref{fig:correction}, both correction methods improve over the original
hypothesis on both datasets, confirming the effectiveness of
target-reference-free correction. More importantly, rubric guidance
consistently yields further gains: MER decreases from 10.43\% to 9.58\% on
Canvas Speech and from 12.94\% to 11.96\% on TED-LIUM 3, while
$\mathrm{SIER}_{\mathrm{lex}}$ decreases from 6.96\% to 5.31\% and from
5.11\% to 3.93\%, respectively. The rubric also repairs more semantic error
mass on both datasets (5.24 vs.\ 3.55 and 4.75 vs.\ 3.63), showing that its
additional benefit comes primarily from more effective repair of
semantically consequential errors.
\vspace{-5pt}
\vspace{-15pt}
\vspace{-5pt}
\subsection{Sensitivity to the Evaluation LLM}
\label{sec:llm-sensitivity}
\vspace{-5pt}
As shown in Table~\ref{tab:llm-sensitivity},
SMART achieves consistently strong agreement with human
judgments across different evaluation LLMs.
Despite variations in precision and recall, all four
models achieve comparable overall classification
performance, suggesting that SMART is relatively
insensitive to the choice of evaluation LLM and
does not rely on a single model for effective
semantic consequence assessment.
\vspace{-10pt}
\begin{table}[h]
\caption{\textbf{Sensitivity of SMART to the evaluation LLM.}
Semantic consequence classification on the same human-annotated segments using different evaluation LLMs.
\textsc{Consequential} is the positive class.
Acc., Prec., Rec., and F1 denote accuracy, precision,
recall, and F1-score, respectively.
Bold indicates the highest value in each column.}
\label{tab:llm-sensitivity}
\centering
\footnotesize
\small
\setlength{\tabcolsep}{2pt}
\begin{tabular*}{\columnwidth}{@{\extracolsep{\fill}}lrrrr@{}}
\hline
\textbf{Evaluation LLM}
& \textbf{Acc. $\uparrow$}
& \textbf{Prec. $\uparrow$}
& \textbf{Rec. $\uparrow$}
& \textbf{F1 $\uparrow$} \\
\hline
GPT-4.1
& 0.895 & 0.750 & 0.800 & 0.774 \\
GPT-5.5
& 0.915 & \textbf{0.868} & 0.733 & 0.795 \\
GPT-5.6-Luna
& 0.895 & 0.714 & \textbf{0.889} & 0.792 \\
DeepSeek-V4-Flash
& \textbf{0.925} & 0.857 & 0.830 & \textbf{0.843} \\
\hline
\end{tabular*}
\end{table}
\vspace{-20pt}

\section{Conclusion}
\vspace{-5pt}
We formulated ASR semantic evaluation as context-aware,
segment-level consequence assessment, enabling recognition
errors to be individually judged and traced to their
semantic impact. To address this task, we proposed \method{},
which combines coherent segment mapping, structured contextual
rubrics, and deterministic aggregation to produce explicit
consequence labels and multi-granularity SIER. Experiments
demonstrate improved agreement with human judgments over direct
LLM evaluation, while results across four evaluation LLMs
suggest that the framework does not rely on a single model.
MER and SIER offer complementary views of lexical accuracy
and semantic impact. A separate target-reference-free rubric
further improves ASR correction using only the target
hypothesis and surrounding ASR hypotheses. These findings
support fine-grained semantic diagnosis and its potential
use in future ASR evaluation and correction.

\section{Compliance with Ethical Standards}
\label{sec:ethics}
\vspace{-5pt}
This study uses the publicly available TED-LIUM 3 corpus
and an existing internal classroom speech corpus.
No new human participants were recruited or recorded,
and no animal experiments were conducted.
The datasets were used in accordance with applicable
data-access and privacy requirements, and the results
are reported in aggregate.
\vspace{-5pt}
\bibliographystyle{IEEEbib}
\begingroup
\fontsize{9pt}{9.5pt}\selectfont
\let\originalthebibliography\thebibliography
\renewcommand{\thebibliography}[1]{%
  \originalthebibliography{#1}%
  \setlength{\itemsep}{0pt}%
  \setlength{\parsep}{0pt}%
}
\bibliography{refs}
\endgroup

\end{document}